\documentclass[numbers,preprint,review,12pt]{elsarticle}

\usepackage{hyperref}
\journal{Neurocomputing}
\usepackage{graphicx}
\usepackage{multirow}
\usepackage{times}
\usepackage{amsmath,amsthm}
\usepackage{amssymb}
\usepackage{amsfonts}
\usepackage{algorithm}
\usepackage{mathrsfs}
\usepackage{booktabs} % For formal tables
\usepackage{color}
\usepackage{hyperref} 
\usepackage{url}
\usepackage{caption} %It's Incompatible for AAAI 
\usepackage{algpseudocode}
\usepackage{pifont}
\usepackage{bm}
\usepackage{comment}
\usepackage{wasysym}
\usepackage{soul}
\usepackage{CJKutf8}

\begin{document}

\begin{frontmatter}



\title{MA-CoNav: A Master-Slave Multi-Agent Framework with Hierarchical Collaboration and Dual-Level Reflection for Long-Horizon Embodied VLN}

\author[swufe]{Ling Luo}
\author[swufe]{Qianqian Bai}
% \author[swufe-a]{Zhongpu Chen}
% \author[swufe-a]{Huaming Du}
% \author[swufe-b]{Yuqian Lei}
% \author[swufe-c]{Ziyun Jiao}
% \cortext[mycorrespondingauthor]{Corresponding author}\ead{zhaoyu@swufe.edu.cn(Yu Zhao);7220323@ustc.edu.cn}
% \author[swufe-a]{Yu Zhao}

% \cortext[mycorrespondingauthor]{Corresponding author}\ead{kougang@swufe.edu.cn}

\address[swufe]{Southwestern University of Finance and Economics, Chengdu, 611130, China}
% \address[swufe-b]{Department of Informatics,Universitat Hamburg, Hamburg, Germany \\}
% \address[swufe-c]{University of Electronic Science and Technology, Chengdu, China \\}
% % \address[zfz-address]{Institute of Artificial Intelligence, Beihang University, Beijing 100191, China \\
% SKLSDE, School of Computer Science, Beihang University, Beijing 100191, China}

%% Abstract
\begin{abstract}
% 视觉语言导航旨在赋予机器人根据复杂语言指令，在陌生环境中进行长程导航的能力。其成功的关键，在于建立高效的“语言理解-视觉感知-具身执行”闭环。现有方法在复杂长距离任务中，常因单智能体认知过载导致感知失真与决策漂移。本文受分布式认知理论启发，提出MA-CoNav（多智能体协同导航框架）。该框架采用“主-子”智能体层级协作架构，将导航任务所需的感知、规划、执行与记忆功能解耦并分布到专业化智能体上。具体而言，主智能体负责全局统筹；子智能体群则分工协作：观察智能体生成环境描述，规划智能体进行任务分解与动态验证，控制执行智能体同步建图与行动，记忆智能体管理结构化经验。此外，框架引入 “局部 - 全局” 双阶反思机制，对全链路导航动作进行动态优化。本文基于 Limo Pro 机器人采集的真实室内数据集开展实证实验，全程未对模型进行任何场景微调。实验结果表明，MA-CoNav在多项指标上全面优于现有的主流视觉语言导航方法。

Vision-Language Navigation (VLN) aims to empower robots with the ability to perform long-horizon navigation in unfamiliar environments based on complex linguistic instructions. Its success critically hinges on establishing an efficient ``language-understanding---visual-perception---embodied-execution'' closed loop. Existing methods often suffer from perceptual distortion and decision drift in complex, long-distance tasks due to the cognitive overload of a single agent. Inspired by distributed cognition theory, this paper proposes MA-CoNav, a Multi-Agent Collaborative Navigation framework. This framework adopts a ``Master-Slave'' hierarchical agent collaboration architecture, decoupling and distributing the perception, planning, execution, and memory functions required for navigation tasks to specialized agents. Specifically, the Master Agent is responsible for global orchestration, while the Subordinate Agent group collaborates through a clear division of labor: an Observation Agent generates environment descriptions, a Planning Agent performs task decomposition and dynamic verification, an Execution Agent handles simultaneous mapping and action, and a Memory Agent manages structured experiences. Furthermore, the framework introduces a ``Local-Global'' dual-stage reflection mechanism to dynamically optimize the entire navigation pipeline. Empirical experiments were conducted using a real-world indoor dataset collected by a Limo Pro robot, with no scene-specific fine-tuning performed on the models throughout the process. The results demonstrate that MA-CoNav comprehensively outperforms existing mainstream VLN methods across multiple metrics.

\end{abstract}

%% Keywords
\begin{keyword}
Vision-and-Language Navigation \sep Multi-Agent Systems \sep Embodied Intelligence \sep  Collaborative Planning




\end{keyword}

\end{frontmatter}

% \linenumbers


\section{Introduction}

In the field of embodied intelligence, Visual-Language Navigation (VLN) \cite{anderson2018vision} stands as one of the core research directions. It aims to empower agents with the ability to autonomously plan paths and complete navigation tasks in unknown or complex environments based on natural language instructions. Its primary objective is to construct an efficient closed loop of "language understanding, visual perception, and embodied execution" by co-optimizing multimodal information processing and action decision-making mechanisms.

Although significant progress has been made in recent years in VLN research based on Large Language Models (LLMs) \cite{zhang2024vision}, agents are prone to cognitive overload when perception, decision-making, planning, and other functions are coupled within a single model \cite{anderson2018vision, zhou2024navgpt, cui2023drivellm} for complex instruction-following tasks in unfamiliar environments. Specific manifestations include: interference among different internal functional modules of the agent, "hallucinations" arising from early minor perceptual biases or spatial misjudgments, difficulty in balancing immediate reactions with long-term planning, ultimately leading to navigation task failure.

To address these bottlenecks, the research frontier is exploring the paradigm of "Distributed Collaborative Intelligence". Distributed cognition theory posits that advanced intelligence is not centralized in a single processing unit but is distributed across multiple specialized functional modules and the interactive media between them. By decoupling complex navigation tasks into sub-functions such as perception, memory, reasoning, and execution, and assigning them to different specialized "cognitive agents," the system can reduce the processing load on individual units. Through information exchange and collaborative verification among agents, more robust and flexible decision-making can be achieved. Recently, studies have proposed parallel collaborative multi-agent frameworks \cite{shridhar2020alfred, wang2023voyager, wei2025rctamp}, which demonstrate stronger task coordination capabilities in dynamic, open worlds through a "multi-round deliberation, consensus planning" model. These works collectively point in one direction: constructing a multi-agent system with clear role division and coordination mechanisms is an effective way to break through the bottlenecks of complex embodied tasks.

However, significant gaps remain in the systematic application of distributed collaborative ideas to visual navigation tasks. Existing research either focuses on single aspects like global mapping or local path planning \cite{rana2023sayplan, chen-etal-2024-mapgpt}, or concentrates on leveraging foundation models to enhance the agent's perception and reasoning capabilities \cite{pan2024langnav, zhou2025learning, zhang2024vision}. There is still a lack of a unified collaborative framework that organically integrates high-level language instruction comprehension, fine-grained environmental perception, hierarchical task planning, real-time action execution, and experiential reflective learning.

Inspired by distributed cognition theory, this paper proposes MA-CoNav, a Multi-Agent Collaborative Navigation framework for vision-and-language navigation. The core innovation of this framework lies in constructing a "master-subordinate" hierarchical agent collaboration model. It decouples the traditionally integrated navigation cognitive functions and allocates them to four specialized sub-agents: The \textit{Observation Agent} generates navigation-oriented environment descriptions through multimodal fusion; the \textit{Task Planning Agent} adopts a hierarchical subtask decomposition strategy and dynamically verifies task completion by incorporating environmental feedback; the \textit{Execution Agent} simultaneously performs incremental map construction and navigation action execution; the \textit{Memory Agent} builds a structured error case library to store and reuse reflection experiences. A master agent is responsible for global task management and inter-agent coordination. Building upon this distributed architecture, we further introduce a "local-global" dual-stage reflection mechanism. Drawing on dual-learning mechanisms, this enables the system to perform immediate correction at the execution level and conduct strategic optimization at the task level, achieving dynamic enhancement throughout the entire navigation pipeline.

To validate the effectiveness of the MA-CoNav framework, this paper conducts empirical experiments using a real-world indoor dataset collected by a Limo Pro robot. It is noteworthy that throughout the entire experimental process, no scene-specific fine-tuning was performed on the models, maximizing the objectivity and generalizability of the experimental results. The results show that MA-CoNav outperforms existing mainstream visual-language navigation methods across several metrics.

In summary, our contributions are as follows:
\begin{itemize}
    \item We propose the MA-CoNav multi-agent framework, which establishes a master-subordinate hierarchical collaboration model to alleviate the cognitive overload of single agents and address complex, long-horizon VLN tasks through division of labor.
    \item We design a multi-role sub-agent system covering the entire pipeline of perception, planning, execution, and memory, realizing a "language-vision-execution" closed loop. Simultaneously, we introduce a "local-global" dual-stage reflection supervision mechanism to efficiently handle motion failures.
    \item Real-world implementation of the framework on a quadruped robot, which autonomously executes a variety of real-time navigation and search tasks in the environment using on-board sensors and computational resources.
\end{itemize}

\section{Related Work}
\subsection{Vision-and-language Navigation}
% 视觉语言导航（Vision-and-Language Navigation, VLN）\cite{anderson2018vision,qi2020reverie,ku2020room}旨在让智能体依据自然语言指令，在复杂环境中自主完成导航任务。近期，大语言模型（LLMs）\cite{chiang2023vicuna,zhu2023chatgpt}凭借其强大的语义与推理能力被引入作为核心决策器，相关研究主要沿两个方向展开。

% 一类工作侧重于将LLMs作为高层任务规划器，旨在将复杂指令解析为可执行的动作序列。例如，DiscussNav \cite{long2024discuss}通过构建多个具备特定角色的智能体进行内部辩论，以生成更可靠的导航决策；NavGPT \cite{zhou2024navgpt}则将视觉观察与历史轨迹编码为文本提示，驱动LLM进行逐步的状态推理与动作生成；VLN-R1框架 \cite{qi2025vln}进一步以第一人称视频流为输入，实现端到端的连续控制。

% 另一类工作则致力于缩小通用LLM与VLN领域之间的差距。例如，NavCoT \cite{lin2025navcot}通过参数高效的领域自适应训练，使LLM能够进行自我引导的链条式思考；KRSP \cite{zhou2024learning}模型则利用知识增强与软提示学习，来提升对局部视觉特征与指令关键词的细粒度对齐能力；也有研究利用递归双向跨模态推理网络（RBCRN），以缓解长序列导航中的历史信息遗忘问题\cite{wu2025recursive}。

% 尽管这些方法提升了指令理解的灵活性，但它们大多仍将环境感知、语言理解、空间推理、路径规划与动作生成等异构认知功能，紧密耦合于单个LLM的前向计算过程中。这种“全能单体”架构在处理复杂、长距离的具身导航任务时，极易引发认知过载。这一根本性瓶颈，揭示了从集中式架构转向分布式协同范式的必要性。

Vision-and-Language Navigation (VLN) \cite{anderson2018vision,qi2020reverie,ku2020room} aims to enable agents to autonomously complete navigation tasks in complex environments based on natural language instructions. Recently, LLMs \cite{chiang2023vicuna,zhu2023chatgpt} have been introduced as core decision-makers due to their powerful semantic understanding and reasoning capabilities, with related research primarily advancing along two directions.

One line of work focuses on employing LLMs as high-level task planners, aiming to parse complex instructions into executable action sequences. For instance, DiscussNav \cite{long2024discuss} constructs multiple specialized agents for internal debate to generate more reliable navigation decisions; NavGPT \cite{zhou2024navgpt} encodes visual observations and historical trajectories into text prompts to drive the LLM for step-by-step state reasoning and action generation; the VLN-R1 framework \cite{qi2025vln} further takes egocentric video streams as input to achieve end-to-end continuous control.

Another line of work strives to bridge the gap between general-purpose LLMs and the VLN domain. For example, NavCoT \cite{lin2025navcot} employs parameter-efficient domain-adaptive tuning to enable LLMs to perform self-guided chain-of-thought reasoning; the KRSP model \cite{zhou2024learning} utilizes knowledge-enhanced reasoning and soft prompt learning to improve fine-grained alignment between local visual features and instruction keywords; other studies employ Recursive Bidirectional Cross-modal Reasoning Networks (RBCRN) to mitigate historical information forgetting in long-horizon navigation \cite{wu2025recursive}.

Although these methods enhance the flexibility of instruction understanding, most still tightly couple heterogeneous cognitive functions—such as environmental perception, language comprehension, spatial reasoning, path planning, and action generation—within the forward computation process of a single LLM. This “all-in-one monolithic” architecture is prone to cognitive overload when handling complex, long-horizon embodied navigation tasks. This fundamental bottleneck underscores the necessity of transitioning from a centralized architecture to a distributed collaborative paradigm.

\subsection{Multi-Agent System}
% 多智能体系统是指由多个具备自主决策与交互能力的智能体，通过特定机制进行协作以解决复杂问题的计算范式\cite{tian2025outlook}。近年来，随着大语言模型（LLM）和视觉语言模型（VLM）~\cite{liu2023visual,chen2024internvl,zhang2024llama}的快速发展，基于大模型的多智能体研究取得了显著进展。\cite{talebirad2023multi}等人的研究表明，在多智能体协作环境中，多个不同角色的智能体协同工作，其综合性能往往优于单智能体。现有前沿工作主要探索通过设计智能体间的对话、辩论或角色扮演等交互机制~\cite{rasal2024llm,shanahan2023role}，以协同解决复杂问题。然而，这些通用框架存在固有缺陷：其一，它们多采用平行协商架构，智能体间通信开销大，显著增加系统复杂性与决策延迟；其二，它们并非为视觉语言导航这类对实时性、空间连续性与环境复杂性有严苛要求的具身任务所设计，缺乏领域定制的协同范式和物理约束建模。

% 为解决上述问题，本文提出的MA-CoNav框架结合分布式认知理论与主从式分层协作架构，针对视觉语言导航任务，明确划分了“全局任务统筹”与“局部专业执行”的层级化过程。

A multi-agent system refers to a computational paradigm in which multiple agents, possessing autonomous decision-making and interaction capabilities, collaborate through specific mechanisms to solve complex problems \cite{tian2025outlook}. In recent years, with the rapid development of large language models (LLMs) \cite{chiang2023vicuna,zhu2023chatgpt} and vision-language models (VLMs) \cite{liu2023visual,chen2024internvl,zhang2024llama}, research on multi-agent systems based on these foundational models has achieved remarkable progress. Studies have shown that in a collaborative multi-agent environment, the collective performance of multiple agents with different roles often surpasses that of a single agent \cite{talebirad2023multi}. State-of-the-art work primarily explores interaction mechanisms such as dialogue, debate, or role-playing among agents \cite{rasal2024llm,shanahan2023role} to collaboratively address complex problems. However, these general frameworks suffer from inherent limitations: firstly, they predominantly adopt a parallel negotiation architecture, where high inter-agent communication overhead significantly increases system complexity and decision latency; secondly, they are not designed for embodied tasks like Visual-Language Navigation (VLN), which have stringent requirements for real-time performance, spatial continuity, and environmental complexity, thus lacking domain-specific collaborative paradigms and physical constraint modeling.

To address the aforementioned issues, the MA-CoNav framework proposed in this paper integrates distributed cognition theory with a master-subordinate hierarchical collaboration architecture. It clearly delineates a hierarchical process of "global task orchestration" and "local specialized execution" specifically for the VLN task.

\section{Methodology}

\begin{figure*}[htb]
    \centering
    \includegraphics[width=1\textwidth]{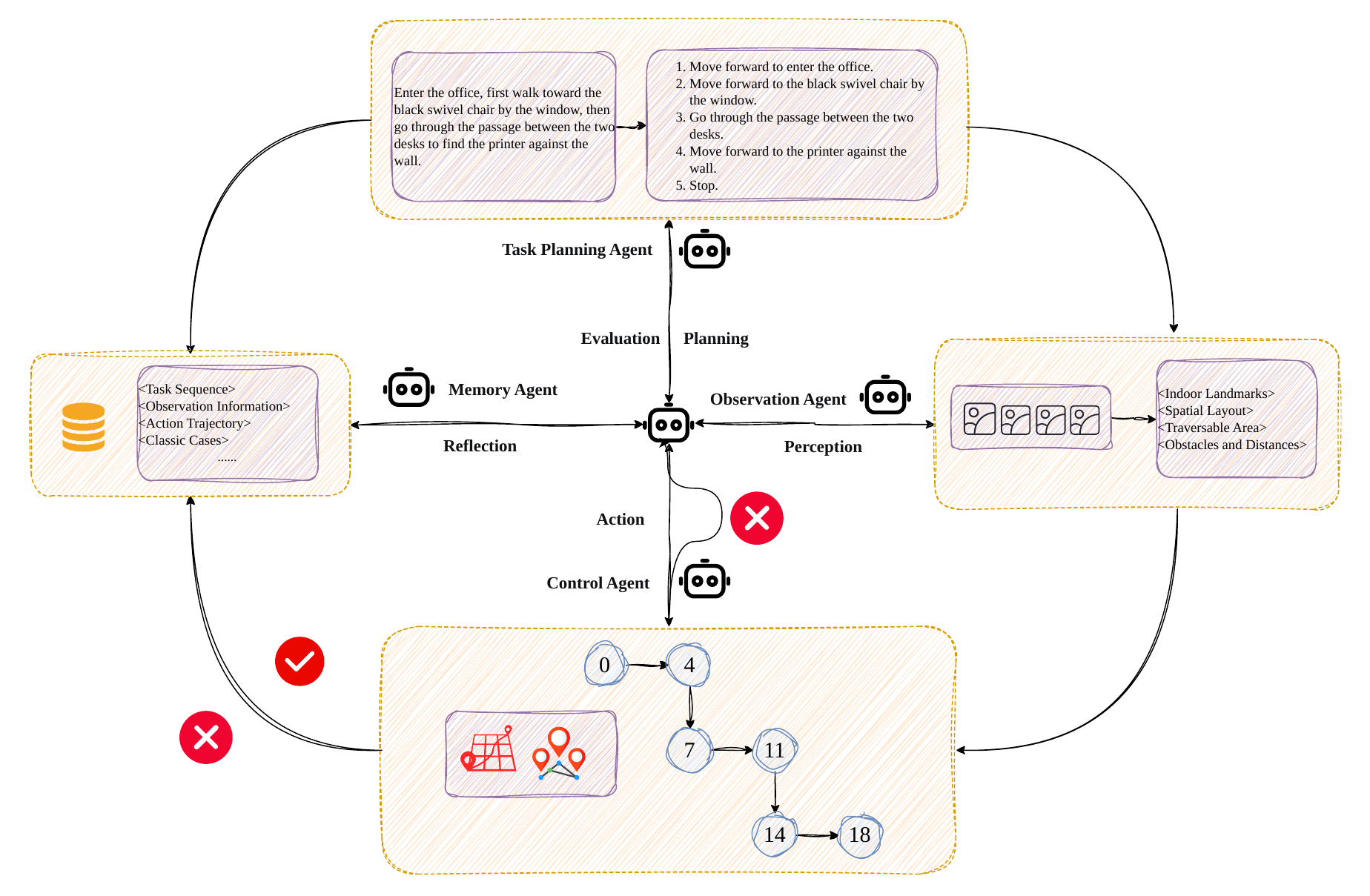} % 调整图片宽度
    \caption{The overall architecture of MA-CoNav.} % 图片标题
    \label{fig:MA-CoNav_Architecture} % 图片标签，用于引用
\end{figure*}

The MA-CoNav framework aims to address the cognitive overload problem faced by traditional methods in complex long-distance navigation tasks through a multi-agent collaboration mechanism, thereby improving the accuracy and robustness of navigation. This section will detail the multi-agent collaborative Vision-and-Language Navigation (VLN) system, including its system composition, collaboration mechanism, and complete workflow.

\subsection{Overall Framework}
% \label{sec:Hippocampal Memory Center}
% \section{整体框架}
% 本文提出MA-CoNav框架，旨在将传统VLN中耦合于单一模型的“全能”认知功能，解耦为由主智能体统一调度的多智能体协作系统。如图2所示，框架采用“1主+4子”的层级结构：

% \begin{itemize}
%     \item \textbf{主智能体（Master Agent）}：作为系统的“指挥官”与“总线”，不直接参与具体的感知或动作生成，而是专注于**任务流控制、信息分发与状态管理**。它维护全局协同循环，根据当前任务状态动态激活相应的子智能体。
%     \item \textbf{子智能体（Sub-agents）}：包括任务规划、观察、控制执行与记忆智能体，作为功能专精的模块，仅在主智能体激活时工作，并向主智能体反馈结构化结果。
% \end{itemize}

% 框架的核心运行机制是一个由主智能体驱动的**“规划-感知-执行-评估”闭环**。主智能体接收自然语言指令后，首先调用任务规划智能体生成子任务序列；随后进入循环调度模式：按需唤醒观察智能体获取环境语义，将语义传递给控制执行智能体进行动作决策，并在动作完成后调用任务规划智能体校验子任务完成状态。全程产生的数据流均由记忆智能体实时归档，形成可追溯的协同工作流。

This paper proposes the MA-CoNav framework, which aims to decouple the omnipotent cognitive functions typically coupled within a single model in traditional Vision-and-Language Navigation (VLN) into a multi-agent collaborative system unified and scheduled by a Master Agent. As illustrated in Fig. \ref{fig:MA-CoNav_Architecture}, the framework adopts a hierarchical structure comprising ``1 Master + 4 Sub-agents'':

\begin{itemize}
    \item \textbf{Master Agent}: Acting as the system's ``commander'' and ``bus,'' the Master Agent does not directly participate in specific perception or action generation. Instead, it focuses on \textbf{task flow control, information distribution, and state management}. It maintains the global coordination loop and dynamically activates the corresponding sub-agents based on the current task state.
    \item \textbf{Sub-agents}: These include the Task Planning, Observation, Control Execution, and Memory Agents. Operating as functionally specialized modules, they work only when activated by the Master Agent and report structured results back to it.
\end{itemize}

The core operating mechanism of the framework is a \textbf{``Plan-Perceive-Act-Evaluate'' closed loop} driven by the Master Agent. Upon receiving a natural language instruction, the Master Agent first invokes the Task Planning Agent to generate a sequence of sub-tasks. Subsequently, it enters a cyclic scheduling mode: activating the Observation Agent on demand to acquire environmental semantics, passing these semantics to the Control Execution Agent for action decision-making, and invoking the Task Planning Agent to verify the sub-task completion status after the action is performed. The data flow generated throughout the process is archived in real-time by the Memory Agent, forming a traceable collaborative workflow.

\subsection{Collaborative Workflow of Specialized Sub-Agents}

\subsubsection{Master Agent: Global Scheduling and Coordination Hub}

% \subsection{主智能体：全局调度与协同中枢}
% 主智能体是MA-CoNav框架的逻辑核心，其本质是一个基于大语言模型（LLM）的状态机控制器。它负责维护系统全局状态 \( \mathcal{S}_{global} \)，并依据当前状态决定下一时刻激活哪一个子智能体（Agent Selection）以及传递何种上下文信息（Message Passing）。

% 主智能体的控制逻辑遵循一个标准化的协同循环，该过程可形式化为策略函数 \( \Pi_{Master} \)：
% \[
% (\text{Agent}_{next}, \text{Context}_{out}) = \Pi_{Master}(I, s_{current}, \text{State}_{prev}, \text{Feedback}_{prev})
% \]
% 其中，\(I\) 为原始指令，\(s_{current}\) 为当前聚焦的子任务。协同循环的具体控制流程如下：

% \begin{enumerate}
%     \item \textbf{初始化与分解阶段}：系统启动时，主智能体处于 \texttt{PLANNING} 状态。它将原始指令 \(I\) 传递给\textbf{任务规划智能体}，接收并存储生成的子任务序列 \(S\)，随后将第一个子任务 \(s_1\) 设为当前焦点，状态流转至 \texttt{PERCEPTION}。
    
%     \item \textbf{感知分发阶段}：在 \texttt{PERCEPTION} 状态，主智能体激活\textbf{观察智能体}。它将当前视觉帧与子任务 \(s_{current}\) 打包发送给观察智能体，要求其返回任务导向的环境描述 \(D_t\)。收到反馈后，主智能体将状态流转至 \texttt{ACTION}。
    
%     \item \textbf{执行控制阶段}：在 \texttt{ACTION} 状态，主智能体激活\textbf{控制执行智能体}。它将环境描述 \(D_t\) 与子任务 \(s_{current}\) 传递给控制端，授权其进行地图更新与动作决策。若动作执行成功，主智能体进入 \texttt{EVALUATION} 状态；若执行失败或遇到阻碍，则触发 \texttt{REFLECTION} 机制（调用记忆智能体）。
    
%     \item \textbf{状态校验阶段}：在 \texttt{EVALUATION} 状态，主智能体再次调用\textbf{任务规划智能体}（处于校验模式），询问“当前子任务是否完成？”。
%     \begin{itemize}
%         \item 若反馈 \texttt{True}：主智能体更新焦点至 \(s_{i+1}\)，重置循环至 \texttt{PERCEPTION}。若所有子任务完成，则终止导航。
%         \item 若反馈 \texttt{False}：保持当前子任务不变，循环回退至 \texttt{PERCEPTION}，继续执行寻找或移动操作。
%     \end{itemize}
% \end{enumerate}

% 通过上述机制，主智能体确立了严格的通信协议，解决了多智能体系统中常见的信息孤岛与时序混乱问题。

The Master Agent serves as the logical core of the MA-CoNav framework. Essentially, it is a state-machine controller based on a Large Language Model (LLM). It is responsible for maintaining the global system state \( \mathcal{S}_{global} \) and, based on the current state, determining which sub-agent to activate next (Agent Selection) and what contextual information to transfer (Message Passing).

The master agent control logic follows a standardized coordination loop, which can be formalized as a policy function \( \Pi_{Master} \):

\begin{equation}
    (\text{Agent}_{next}, \text{Context}_{out}) = \Pi_{Master}(I, s_{current}, \text{State}_{prev}, \text{Feedback}_{prev})
\end{equation}

Where \(I\) is the raw instruction, and \(s_{current}\) is the sub-task currently in focus. The specific control flow of the coordination loop is as follows:

\begin{enumerate}
    \item \textbf{Initialization and Decomposition Phase}: Upon system startup, the Master Agent is in the \texttt{PLANNING} state. It transmits the raw instruction \(I\) to the \textbf{Task Planning Agent}, receives and stores the generated sub-task sequence \(S\), sets the first sub-task \(s_1\) as the current focus, and transitions the state to \texttt{PERCEPTION}.
    
    \item \textbf{Perception Distribution Phase}: In the \texttt{PERCEPTION} state, the Master Agent activates the \textbf{Observation Agent}. It packages the current visual frame and the sub-task \(s_{current}\) to send to the Observation Agent, requesting a task-oriented environment description \(D_t\). Upon receiving the feedback, the Master Agent transitions the state to \texttt{ACTION}.
    
    \item \textbf{Execution Control Phase}: In the \texttt{ACTION} state, the Master Agent activates the \textbf{Control Execution Agent}. It passes the environment description \(D_t\) and the sub-task \(s_{current}\) to the control unit, authorizing it to perform map updates and action decision-making. If the action is executed successfully, the Master Agent enters the \texttt{EVALUATION} state; if execution fails or obstacles are encountered, the \texttt{REFLECTION} mechanism is triggered (invoking the Memory Agent).
    
    \item \textbf{State Verification Phase}: In the \texttt{EVALUATION} state, the Master Agent reinvokes the \textbf{Task Planning Agent} (in verification mode) to query, ``Is the current sub-task completed?''.
    \begin{itemize}
        \item If the feedback is \texttt{True}: The Master Agent updates the focus to \(s_{i+1}\) and resets the loop to \texttt{PERCEPTION}. If all sub-tasks are completed, navigation terminates.
        \item If the feedback is \texttt{False}: The current sub-task remains unchanged, and the loop rolls back to \texttt{PERCEPTION} to continue execution of search or movement operations.
    \end{itemize}
\end{enumerate}

Through the aforementioned mechanism, the Master Agent establishes a strict communication protocol, effectively addressing the common issues of information silos and temporal disorder in multi-agent systems.

\subsubsection{Task Planning Agent: Hierarchical Decomposer and Dynamic Verifier}

This agent functions as both the "Instruction Interpreter" and the "Operational Orchestrator" of high-level directives. Its core architecture facilitates the structured decomposition of complex navigation instructions and the dynamic verification of sub-task completion based on multi-modal environmental feedback.

\paragraph{Hierarchical Decomposition.} Upon receiving the raw natural language instruction $N$ from the master agent, the planning agent does not directly map it to low-level control primitives. Instead, it leverages the semantic comprehension and commonsense reasoning capabilities of Large Language Models (LLMs) to recursively parse $N$ into a sequence of logical sub-tasks $S=\{s_1, s_2, \dots, s_k\}$, which are grounded in spatial and functional semantics. This process is formally defined as:
\begin{equation}
S = G_{\text{LLM}}(N \mid \Theta_{\text{plan}})
\end{equation}
where the prompt $\Theta_{\text{plan}}$ defines the decomposition logic and structural constraints. The decomposition adheres to principles of spatial and logical coherence. For instance, an instruction such as ``walk to the black swivel chair by the window, pass through the aisle between two desks, and find the printer against the wall'' is transformed into a landmark-centric sequence: $\{\text{Office} \rightarrow \text{Swivel Chair} \rightarrow \text{Aisle} \rightarrow \text{Printer}\}$.

\paragraph{Dynamic State Verification.} To ensure robust execution, the agent continuously monitors the environmental descriptions $O_t$ from the observation module and the execution history $H_t$ from the controller, conditioned on the context retrieved from the memory module. At each decision step $t$, it evaluates the semantic completion of the active sub-task $s_i$ relative to the current observation $O_t$ and historical trajectory $H_t$. This verification process is formulated as a state evaluation function:
\begin{equation}
\phi_t = V(s_i, O_t, H_t)
\end{equation}
When the completion progress $\phi_t$ exceeds a predefined threshold $\tau$, the current sub-task is deemed semantically satisfied. Consequently, the planning agent shifts the system's focus to the subsequent sub-task $s_{i+1}$, enabling the incremental realization of complex, long-horizon instructions.

\subsubsection{Observation Agent: Navigation-Oriented Environmental Describer}

The proposed agent serves as the system's ``visual eyes'' and ``on-site interpreter,'' responsible for distilling the first-person visual stream into structured semantic descriptions aligned with the current mission. Unlike generic scene captioning, this agent executes a \textbf{task-driven attention mechanism}. The underlying process is formally partitioned into two stages: \textit{Omnidirectional Multi-view Perception} and \textit{Task-oriented Description Generation}.

Specifically, at each time step $t$, the agent captures raw image inputs from four cardinal directions, denoted as $\{I_{t,1}, I_{t,2}, I_{t,3}, I_{t,4}\}$. During the \textbf{Multi-view Perception} stage, a Multimodal Large Language Model (MLLM) is employed to extract a primary perceptual tuple $P_{t,i}$ for each view $i$:
\begin{equation}
P_{t,i} = (O_{t,i}, L_{t,i}, T_{t,i}, M_{t,i})
\end{equation}
where $O_{t,i}$ denotes a set of obstacles characterized by their categories, pixel-level coordinates, and estimated depths; $L_{t,i}$ represents potential navigational landmarks (e.g., ``door,'' ``desk,'' ``printer'') with their relative bearings; $T_{t,i}$ is the traversability segmentation mask (walkable area); and $M_{t,i}$ provides a concise global context phrase to encapsulate the scene's high-level semantics.

Subsequently, the agent aggregates the perceptual ensemble $P_t = \{P_{t,1}, P_{t,2}, P_{t,3}, P_{t,4}\}$ and integrates these signals with the current sub-task $s_{\text{current}}$. This reasoning and distillation process is governed by a core observation function $F_{\text{Obs}}$:
\begin{equation}
D_t = F_{\text{Obs}}(P_t, s_{\text{current}}; \Theta_{\text{navigation}})
\end{equation}
Guided by the navigation-specific prompt template $\Theta_{\text{navigation}}$, this function performs information distillation and spatial reasoning across the following four dimensions:

\begin{itemize}
    \item \textbf{Semantic Salience Filtering}: Prioritizing objects and landmarks in $P_t$ that exhibit high semantic relevance to $s_{\text{current}}$ (e.g., accentuating office equipment when the goal is ``finding a printer'').
    \item \textbf{Traversability Assessment}: Synthesizing $T_{t,i}$ across all views to identify the optimal navigation path and articulating its topological relationship with key anchors to facilitate downstream trajectory planning.
    \item \textbf{Target Landmark Localization}: Detecting task-critical objects and estimating their relative poses (azimuth and distance) via multi-view geometric cues (e.g., ``a blue bin is situated 1.5m to the front-left'').
    \item \textbf{Spatial Relationship Modeling}: Explicitly encoding the relative dependencies between entities and the agent, transforming multi-view primitives into a structured linguistic representation that explicitly models environmental topology.
\end{itemize}

Through this hierarchical processing, the output $D_t$ effectively abstracts away redundant visual noise, providing the subsequent path planning and control modules with mission-critical, spatially-aware insights.

\subsubsection{Control Execution Agent: The Unification of Map Building and Action Execution}
% 该智能体是系统的“现场制图师”与“执行手足”，核心职责是在未知环境中同步执行增量式地图构建与实时动作决策。它将感知信息转化为一个包含几何与拓扑结构的内部环境表示，并基于此表示选择具体的导航动作，实现“建图-执行”的闭环。

% 智能体维护一个动态更新的局部环境地图 \( \mathcal{M}_t = (\mathcal{G}_t, \mathcal{T}_t) \)，其中 \( \mathcal{G}_t \) 为几何坐标地图，\( \mathcal{T}_t \) 为拓扑地图。

The agent functions as the system's \textbf{in-situ cartographer} and \textbf{action effector}. Its primary responsibility is the concurrent execution of incremental map construction and real-time decision-making within unexplored environments. By transforming raw sensory data into an internal environment representation that encapsulates both geometric and topological structures, the agent enables a robust ``mapping-execution'' closed-loop. Specifically, the agent maintains a dynamically updated local map $M_t = (\mathscr{G}_t, \mathscr{T}_t)$, where $\mathscr{G}_t$ represents the \textit{geometric coordinate map} and $\mathscr{T}_t$ denotes the \textit{topological graph}.

% \subsubsubsection{Geometric Coordinate Map}
The geometric map $\mathscr{G}_t$ is constructed incrementally based on robot odometry. At any time step $t$, the robot tracks its current pose $p_t = (x_t, y_t, \theta_t)$, where $(x_t, y_t)$ are the coordinates in the global frame (initialized at the origin $(0,0)$) and $\theta_t$ is the global heading. To streamline exploratory heuristics, the agent computes candidate navigation waypoints in four quadrature directions (Front, Right, Back, Left). Given a constant unit step size $\delta$, the set of candidate points $C_t$ is defined as:
\begin{equation}
C_t = \{c_t^i \mid i \in \{0,1,2,3\}\}
\end{equation}
\begin{equation}
c_t^i = \left(x_t + \delta \cdot \cos\left(\theta_t - i \cdot \frac{\pi}{2}\right),\ y_t + \delta \cdot \sin\left(\theta_t - i \cdot \frac{\pi}{2}\right)\right)
\end{equation}
Each candidate $c_t^i$ is assigned a unique identifier $v_k$ and archived in the vertex set $V_t$. The geometric map is thus defined as the union of all traversed poses and identified nodes: $\mathscr{G}_t = \{p_0, p_1, \dots, p_t\} \cup V_t$. 

\textit{Example}: For a current pose $p_t = (2, 1, 90^\circ)$ and $\delta = 1$, the quadrature candidates are: Front $(2, 2)$, Right $(3, 1)$, Back $(2, 0)$, and Left $(1, 1)$. These points are registered as nodes $v_5, v_6, v_7, v_8$ within the map structure.

% \subsubsubsection{Topological Map}
The topological map $\mathscr{T}_t = (V_t, E_t)$ abstracts environmental connectivity as a graph, where $V_t$ is the set of nodes (corresponding to key locations or candidates) and $E_t \subseteq V_t \times V_t$ is the edge set representing traversable paths. When the agent moves from the current node $v_c$ to or confirms the reachability of a candidate $v_i$, the topological structure is updated:
\begin{equation}
V_t \leftarrow V_t \cup \{v_i\}, \quad E_t \leftarrow E_t \cup \{(v_c, v_i)\}
\end{equation}
This structure is implemented via an adjacency dictionary (e.g., $\text{Adj}[v_c] = [\dots, v_i]$), facilitating efficient connectivity queries and pathfinding.

% \subsubsubsection{Multi-source Informed Real-time Decision Making}
At each decision step $t$, the agent operates as a policy function $\pi$, synthesizing heterogeneous information to output a low-level action $a_t$:
\begin{equation}
a_t = \pi(s_{\text{current}}, \mathcal{D}_t, \mathcal{M}_t, \mathcal{H}_t)
\end{equation}
where $s_{\text{current}}$ is the sub-task issued by the high-level task planner, $\mathcal{D}_t$ is the task-oriented environment description provided by the observation module, $\mathcal{M}_t$ represents the instantaneous state of the internal map, and $\mathcal{H}_t$ denotes the history of trajectories and prior decisions.

The action space $\mathcal{A}$ comprises discretized \textbf{navigational primitives}: $\mathcal{A} = \{\text{MoveForward, TurnRight } 90^\circ, \text{TurnLeft } 90^\circ, \text{Stop}\}$. This design balances control simplicity with environmental maneuverability. The decision process $\pi$ first filters candidates from $C_t$ that satisfy the semantic requirements of $s_{\text{current}}$ and topological reachability, then employs pathfinding heuristics (e.g., shortest path on $\mathscr{T}_t$) to generate the final execution command.

\subsubsection{Memory Agent: Historical Storage and Experience Management}

The Memory Agent serves as the system's \textbf{Knowledge Custodian}. Its primary responsibility lies in the synchronous archiving and structuring of all historical traces from each navigation mission. Beyond mere logging, it distills high-level navigational insights from raw trajectories, providing a foundational substrate for the continual learning of the multi-agent system.

During mission execution, the Memory Agent captures raw state-action pairs across all time steps. For a mission of duration $T$, the constructed global history $H$ is defined as a time-indexed ensemble:
\begin{equation}
H = \{(t, s_t, a_t, O_t, M_t) \mid t = 0, 1, \dots, T\}
\end{equation}
where $s_t = (x_t, y_t, \theta_t)$ represents the agent's pose at time $t$, $a_t \in \mathscr{A}$ denotes the executed action, $O_t$ signifies the sensory observations, and $M_t = (\mathscr{G}_t, \mathscr{T}_t)$ represents the instantaneous map maintained by the execution agent. This history $H$, alongside mission metadata (e.g., unique identifier $\text{ID}$, timestamp, and mission instruction $I$), is encapsulated into a high-fidelity trajectory record for retrospective analysis.

Upon the completion of a navigation task, the Memory Agent performs a \textbf{post-hoc analysis} to extract critical learning points. Specifically, it identifies representative failures—such as the perceptual misidentification of transparent obstacles (e.g., glass doors) or motion deadlocks in narrow corridors—and encodes them into a structured quadruple containing scene features, erroneous actions, failure attributions, and corrective outcomes. Each experience entry $e_k$ is formally represented as:
\begin{equation}
e_k = \langle \text{ID}_k, F_k, T_k, I_k \rangle
\end{equation}
where $F_k$ is a feature vector encoding scene semantics (including instructions and key landmarks), $T_k = (P, A, O)$ provides the complete trajectory context, and $I_k$ represents the \textbf{Reflective Knowledge Tuple}:
\begin{equation}
I_k = (F_{\text{err}}, a_{\text{err}}, \text{cause}, a_{\text{corr}})
\end{equation}
Here, $F_{\text{err}}$ characterizes the failure-prone scene, $a_{\text{err}}$ is the suboptimal action taken, $\text{cause}$ denotes the inferred root cause, and $a_{\text{corr}}$ specifies the validated corrective strategy. These entries are indexed within a global memory bank $\mathcal{M}$ to facilitate efficient semantic retrieval.

To support experience reuse in novel environments, the agent extracts a query feature $F_q$ from the current observation $O_t$ and map $M_t$. The system then retrieves the most relevant historical cases from the experience bank $E$. This retrieval process is formulated as a \textbf{similarity maximization} problem:
\begin{equation}
e^* = \underset{e_k \in E}{\arg\max} \, \text{Sim}\left(\Psi(F_q), \Psi(F_k)\right)
\end{equation}
where $\Psi(\cdot)$ denotes a feature encoding function that maps raw characteristics into a unified semantic embedding space, and $\text{Sim}(\cdot, \cdot)$ is a similarity metric. The knowledge $K^*$ embedded within the retrieved experience $e^*$ is then leveraged by the task planner or the master agent to adapt current policies or serve as a \textit{proactive risk alert}, thereby closing the loop of continual learning.

% \subsection{Local-Global Two-Stage Reflection Mechanism}

\subsection{Local-Global Dual-Stage Reflection Mechanism}

To elevate passive error correction into proactive learning, the MA-CoNav framework introduces a \textit{Dual-Stage Reflective Mechanism}. This mechanism is designed to establish a complete cognitive closed-loop, ranging from instantaneous online correction to offline experience consolidation. While the \textbf{Local Reflection} focuses on real-time action verification and fine-tuning during sub-task execution to ensure single-step reliability, the \textbf{Global Reflection} performs deep trajectory analysis and knowledge distillation upon task completion. These two stages are bridged by the Memory Agent, forming a tightly coupled reinforcement learning loop with bidirectional information flow.

Local reflection is a lightweight, high-frequency online process. Its core function is to rapidly evaluate the rationality and safety of a decision immediately before or after action execution, aiming to intercept low-level errors caused by transient perceptual noise or model limitations. Driven by the Control-Execution Agent, local reflection is triggered at each decision step $t$. When the agent generates a candidate action $a_t$ based on the internal map $M_t$ and the current sub-task $s_{\text{current}}$, it invokes a rapid verification function $R_{\text{local}}$ rather than executing it immediately:

\begin{equation}
\text{flag}_{\text{local}} = R_{\text{local}}(a_t, O_t, M_t, s_{\text{current}})
\end{equation}

This function determines whether action $a_t$ fundamentally conflicts with the current environmental description $O_t$ provided by the Observation Agent. For instance, if $O_t$ explicitly indicates an "impassable obstacle 1 meter ahead" while the planner proposes a "Move Forward" action, an inconsistency flag is raised. Simultaneously, the abstract features $F_t$ of the current scene are matched against high-risk experiences in the Memory Agent’s case library. If the similarity exceeds a predefined threshold $\tau_{\text{risk}}$, i.e., $\text{Sim}(\Psi(F_t), \Psi(F_{\text{err}})) > \tau_{\text{risk}}$, the situation is flagged as a potential risk scenario.

If $\text{flag}_{\text{local}}$ indicates a conflict or high risk, the Control-Execution Agent initiates a \textit{micro-planning} process: it vetoes the original action $a_t$ and regenerates an alternative action $a_t'$ based on current constraints (e.g., changing "Move Forward" to a "Turn Right 90°" exploratory observation). If the evaluation passes, $a_t$ is executed as planned. Post-execution, the agent performs an instantaneous comparison between the outcome and the expectation. In the event of a severe mismatch (e.g., expected movement but sensors detect a collision), a backtrack is triggered to undo the action and re-decide. All local reflection events—including triggered verifications, adjustments, and outcomes—are synchronously logged into the task history $H$.

Global reflection is activated upon the completion (success or failure) of a navigation task. Coordinated by the Lead Agent, it organizes a "post-mortem" review among sub-agents to scrutinize the execution process from a macro perspective. Taking the complete task history $H$ stored by the Memory Agent and the original instruction $I$ as inputs, the process first segments the continuous history $H$ into meaningful episodes $\{H_1, H_2, \dots, H_m\}$ based on sub-task completion points, prolonged stagnations, or frequent action oscillations. 

Subsequently, for inefficient or failed segments identified during evaluation, the mechanism infers root causes by analyzing the internal states of the agents during those intervals. For example, repeated failures at a glass door may be attributed to a systematic misidentification of visual features rather than a stochastic perceptual error. These analytical results—specifically successful patterns and failure cases—are encapsulated into the previously defined four-element experience tuples $e = \langle D, F, T, I \rangle$. The distilled experience $e$ is then submitted to the Memory Agent for storage ($M \leftarrow M \cup \{e\}$) and index updating.

\section{Experiment}
% 本节将通过全面的现实世界实验评估所提出的MA-CoNav框架。我们的评估旨在回答三个关键问题：
% \begin{enumerate}
%     \item MA-CoNav在真实世界复杂导航任务中的适用性程度？
%     \item 框架中各核心模块（多智能体协同、反思机制）对性能的贡献程度？
%     \item 局部-全局双阶反思机制对导航任务的提升程度？
% \end{enumerate}

This section comprehensively evaluates the proposed MA-CoNav framework through extensive real-world experiments. Our evaluation aims to answer three key questions:
\begin{enumerate}
    \item To what extent is MA-CoNav applicable to complex navigation tasks in the real world?
    \item To what extent do each core module (multi-agent collaboration, reflection mechanism) in the framework contribute to performance?
    \item To what extent does the local-global dual-stage reflection mechanism improve navigation tasks?
\end{enumerate}

% \subsection{Experimental Setting}

% 为全面评估框架性能，设计了基于长距离复合指令的导航任务。该任务要求机器人根据一条自然语言指令，在未知室内环境中顺序定位并访问多个目标地点或物体（例如：“进入办公室，先走向靠窗的黑色转椅，再穿过两张办公桌之间的通道，找到靠墙的打印机”）。相较于单目标导航，此类任务对系统的指令理解、长程规划、序贯执行及环境适应能力提出更高要求。

% \subsubsection{实验环境与平台}
% 将实验组织为4个场景：办公区、实验室、家居环境（包含厨房、卧室、客厅）以及健身房，这些环境包含各种物体类别，如计算机、椅子、微波炉、咖啡机和盆栽植物与功能区域。用于评估MA-CoNav的对不同空间布局和语义的泛化性。所有实验均使用松灵 Limo PRO 移动机器人\footnote{\url{https://iqr-robot.com/product/agilex-limo/}}作为物理智能体，如图5所示。它配备了奥比中光 DaBai 相机用于视觉感知，以30 Hz的频率输出彩色和深度帧，每个帧的分辨率为640×480。框架中除智能体依赖的大型语言模型外，其余所有模块均在机器人机载运行。大型语言模型（GPT-4-Turbo、GPT-5.2 Pro）通过云API访问。整个软件系统在ROS 2 Foxy Fitzroy 中间件上构建与运行。
% \subsubsection{对比方法}
% 将MA-CoNav与与单智能体架构\cite{zhou2024navgpt,chen2024mapgpt}、多智能体架构\cite{brienza2024multi,wei2025rctamp}在相同现实场景中进行比较。值得注意的是，上述基线原实验多基于仿真环境，我们将其核心算法迁移至本实验的真实机器人平台，并保持其感知与执行接口与我们的实验设置一致，确保性能差异主要源于算法架构本身。

% \subsubsection{评估指标}
% 采用视觉语言导航任务评估中被广泛认可的的6个定量指标\cite{krantz2022instance}进行评估。导航路径长度（NL）：从起点到完成所有目标访问的步长；目标距离误差（NE）：任务结束时智能体与最终目标物体的直线距离；任务成功率（SR）：成功按指令顺序完成所有目标访问的任务占比；先知成功率（OSR）：允许智能体首次到达任一目标 1 米范围时强制停止，仅评估路径规划与目标定位能力；路径长度加权成功率（SPL）\cite{anderson2018evaluation}在成功率基础上引入路径效率的惩罚，其计算公式为：
% \[
% \text{SPL} = \frac{1}{N} \sum_{i=1}^{N} S_i \times \frac{L_i^*}{\max(L_i, L_i^*)}
% \]
% 其中，$S_i$ 表示第$i$个任务是否成功（1或0），$L_i$ 为实际行驶路径长度，$L_i^*$ 为该任务的理论最短路径长度；关键点准确率（KPA）：统计智能体在指令关键子任务（如 “穿过办公桌通道”“定位黑色转椅”）上做出正确决策的占比。每条指令在相同初始条件下均重复测试 5 轮，避免偶然因素影响。

\subsection{Experimental Setting}

To comprehensively evaluate the performance of the framework, a navigation task based on long-distance composite instructions is designed. This task requires the robot to sequentially locate and visit multiple target locations or objects in an unknown indoor environment according to a single natural language instruction (e.g.: "Enter the office, first walk to the black swivel chair by the window, then pass through the passage between two office desks, and find the printer against the wall"). Compared with single-target navigation, such tasks impose higher requirements on the system's instruction understanding, long-range planning, sequential execution, and environmental adaptability.

\subsubsection{Experimental Environment and Platform}
The experiments are organized into 4 scenarios: office area, laboratory, home environment (including kitchen, bedroom, living room), and gym. These environments contain various object categories (e.g., computers, chairs, microwave ovens, coffee machines, potted plants) and functional areas, which are used to evaluate the generalization ability of MA-CoNav to different spatial layouts and semantics. All experiments use the AgileX Limo PRO mobile robot\footnote{\url{https://iqr-robot.com/product/agilex-limo/}} as the physical agent, as shown in Figure 5. It is equipped with the ORBBEC DaBai camera for visual perception, outputting color and depth frames at a frequency of 30 Hz with a resolution of 640×480 per frame. Except for the large language model relied on by the agent, all other modules in the framework run on the robot's on-board system. The large language model (GPT-4-Turbo,GPT-5.2 Pro) is accessed via cloud API. The entire software system is built and run on the ROS 2 Foxy Fitzroy middleware.

\subsubsection{Comparative Methods}
MA-CoNav is compared with single-agent architectures\cite{zhou2024navgpt,chen2024mapgpt} and multi-agent architectures\cite{brienza2024multi,wei2025rctamp} in the same real-world scenarios. It is worth noting that the original experiments of the above baselines are mostly based on simulated environments; we migrate their core algorithms to the real robot platform of this experiment, and keep their perception and execution interfaces consistent with our experimental settings to ensure that the performance differences mainly stem from the algorithm architecture itself.

\subsubsection{Evaluation Metrics}
Six widely recognized quantitative metrics\cite{krantz2022instance} in the evaluation of vision-language navigation tasks are adopted. Navigation Path Length (NL): the number of steps from the starting point to the completion of all target visits; Nearness Error (NE): the straight-line distance between the agent and the final target object at the end of the task; Success Rate (SR): the proportion of tasks that successfully complete all target visits in the order of the instruction; Oracle Success Rate (OSR): allowing the agent to stop forcibly when it first reaches within 1 meter of any target, only evaluating path planning and target localization capabilities; Success weighted by Path Length (SPL)\cite{anderson2018evaluation} introduces a penalty for path efficiency on the basis of success rate, and its calculation formula is:
\[
\text{SPL} = \frac{1}{N} \sum_{i=1}^{N} S_i \times \frac{L_i^*}{\max(L_i, L_i^*)}
\]
where $S_i$ indicates whether the $i$-th task is successful (1 or 0), $L_i$ is the actual traveled path length, and $L_i^*$ is the theoretical shortest path length of the task; Key Point Accuracy (KPA): the proportion of correct decisions made by the agent on key subtasks of the instruction (e.g., "pass through the office desk passage", "locate the black swivel chair"). Each instruction is repeatedly tested 5 times under the same initial conditions to avoid the influence of random factors.

% \begin{figure}[htbp]
%     \centering
%     \includegraphics[width=0.3\textwidth]{images/limo.png} % 调整图片宽度
%     \caption{AgileX Limo PRO mobile robot.} % 图片标题
%     \label{fig:limo} % 图片标签，用于引用
% \end{figure}

\subsection{Experimental Results}
\label{sec:experimental_results}

This section presents the experimental results of complex navigation tasks in the real world, aiming to verify the effectiveness of the proposed architecture. All evaluation tasks are based on 50 complex natural language instructions generated by the Multimodal Large Language Model (MLLM) from panoramic views of real environments, where each instruction contains three or more target objects. As shown in Table~\ref{tab:main_comparison}, MA-CoNav significantly outperforms single-agent baselines (NavGPT\cite{zhou2024navgpt}, MapGPT\cite{chen2024mapgpt}) and state-of-the-art multi-agent baselines (CoELA\cite{brienza2024multi}, RCTAMP\cite{wei2025rctamp}) on all metrics.

Specifically, the success rate (SR) of the two single-agent baselines in complex real-world tasks is $0\%$, which intuitively reflects the severe "cognitive overload" problem of traditional coupled architectures when faced with long instructions and multiple objectives, as well as their lack of adaptability in real-world environments. In contrast, MA-CoNav achieves an SR of $25.6\%$, significantly higher than CoELA ($2.8\%$) and RCTAMP ($8.4\%$), indicating that this architecture, through its "master-sub" two-level structure, is more adept at handling long-horizon, multi-objective composite instructions. In terms of the oracle success rate (OSR), which reflects pure path discovery capability, MA-CoNav reaches $37.6\%$, also far exceeding other baselines. This suggests that its collaborative mechanism enables more effective environment exploration and target localization. 

However, the navigation length (NL) of MA-CoNav is $33.56$ steps, higher than other methods. This is attributed to the limited reasoning ability of large language models regarding spatial relationships, which often triggers local reflections in the control execution agent. While this mechanism may lead to more frequent path adjustments, it also successfully intercepts decisions that could lead to failure, ultimately contributing to a higher task success rate. Furthermore, MA-CoNav achieves a significant lead in key point accuracy (KPA) at $69.89\%$, validating the enhanced framework's capability for fine-grained understanding and reliable execution of complex instructions.

\begin{table*}[!t]
    \centering
    \caption{Comparison of MA-CoNav with State-of-the-Art Methods}
    \label{tab:main_comparison}
    \resizebox{\linewidth}{!}{%
\begin{tabular}{l l c c c c c c}
\toprule
\textbf{Category} & \textbf{Method} & \textbf{NL (Steps)}$\downarrow$ & \textbf{NE (Meters)}$\downarrow$ & \textbf{SR (\%)}$\uparrow$ & \textbf{OSR (\%)}$\uparrow$ & \textbf{SPL (\%)}$\uparrow$ & \textbf{KPA (\%)}$\uparrow$ \\
\midrule
% 替换原 multirow{2}{*}{Single-Agent}，第一行显示文字，第二行留空
\textbf{Single-Agent} & NavGPT\cite{zhou2024navgpt} & 18.67 & 6.78 & 0 & 2.00 & 0 & 14.14 \\
                      & MapGPT\cite{chen2024mapgpt} & 11.21 & 5.97 & 0 & 4.80 & 0 & 21.82 \\
\midrule
% 替换原 multirow{3}{*}{Multi-Agent}，第一行显示文字，后两行留空
\textbf{Multi-Agent}  & CoELA\cite{brienza2024multi} & 34.78 & 5.71 & 2.80 & 5.20 & 1.15 & 32.01 \\
                      & RCTAMP\cite{wei2025rctamp} & 38.59 & 4.27 & 8.40 & 12.80 & 5.04 & 34.54 \\
                      & \textbf{MA-CoNav (Ours)} & \textbf{33.56} & \textbf{2.93} & \textbf{25.60} & \textbf{37.60} & \textbf{15.62} & \textbf{69.89} \\
\bottomrule
\end{tabular}
  }
\end{table*}

In addition, Table~\ref{tab:llm_config} evaluates the performance of different Large Language Models (LLMs) within the MA-CoNav framework. It is found that the method using only MLLM generally achieves lower performance than the method combining MLLM with LLM. This indicates that without the assistance of external modules, MLLMs struggle to effectively integrate current observed images, navigation texts and historical information to navigate to target positions accurately.

\begin{table}[!t]
  \centering
  \caption{Comparison of Different LLM Configurations under the MA-CoNav Framework}
    \label{tab:llm_config}
  \resizebox{\linewidth}{!}{%
  \begin{tabular}{lcccccc}
    \toprule
    \textbf{Model} & \textbf{NL (Steps)}$\downarrow$ & \textbf{NE (Meters)}$\downarrow$ & \textbf{SR (\%)}$\uparrow$ & \textbf{OSR (\%)}$\uparrow$ & \textbf{SPL (\%)}$\uparrow$ & \textbf{KPA (\%)}$\uparrow$ \\
    \midrule
    MLLM & 32.17 & 4.05 & 17.60 & 32.80 & 10.56 & 51.92 \\
    MLLM + LLM & 33.56 & 2.93 & 25.60 & 37.60 & 15.62 & 69.89 \\
    \bottomrule
  \end{tabular}
  }
\end{table}

\subsection{Ablation Study}

\subsubsection{Framework-level Ablation Experiments}
To analyze the necessity and specific contributions of each core module in the MA-CoNav framework, we designed and conducted a series of ablation experiments based on the same hardware platform and environmental settings as the main experiments. The contribution of each core agent to the overall performance was evaluated by sequentially removing it. After removing a module, each variant only adopted the simplest direct logic to replace its function (e.g., after removing the task planning agent, the original instructions were directly input), so as to isolate and evaluate the contribution of the module.

Table~\ref{tab:framework_ablation} presents the complete results of the ablation study, clearly demonstrating that all agent modules are essential for achieving the system's optimal performance. The Control Agent has the most significant impact; removing this component renders the framework completely incapable of navigation (SR=0\%). This is followed by the removal of the Task Planning Agent, which causes a sharp drop in SR to 5.2\% and results in the lowest path efficiency (SPL), indicating that it is the core component for managing complex task logic and preventing error accumulation. Removing the Master Agent leads to a breakdown in coordination, with SR decreasing by approximately 6 percentage points and a notable reduction in KPA. The absence of the Observation Agent results in a lack of descriptive capability and insufficient spatial awareness, leading to an increased navigation error (NE) and a lower KPA. Finally, although removing the Memory Agent or the Reflection Mechanism does not paralyze the system, both lead to an observable decline in SR of about 8 to 10 percentage points. This demonstrates that these two components play a clear and sustained role in improving success rate, path efficiency, and decision quality (KPA). The ablation study confirms that every component of MA-CoNav contributes significantly to the final performance.

\begin{table}[htbp]
  \centering
  \caption{Ablation Experiment Analysis of MA-CoNav Core Components}
  \label{tab:framework_ablation}
  \resizebox{\linewidth}{!}{%
  \begin{tabular}{lcccccc}
    \toprule
    Method & NL (Steps) $\downarrow$ & NE (Meters) $\downarrow$ & SR (\%) $\uparrow$ & OSR (\%) $\uparrow$ & SPL (\%) $\uparrow$ & KPA (\%) $\uparrow$ \\
    \midrule
    w/o Main Agent & 33.64 & 4.25 & 19.60 & 24.80 & 11.96 & 44.95 \\
    w/o Task Planning Agent & 18.31 & 4.92 & 5.20 & 9.20 & 3.02 & 28.18 \\
    w/o Observation Agent & 39.57 & 5.63 & 8.40 & 13.20 & 4.62 & 30.81 \\
    w/o Control Execution Agent & 12.46 & 8.71 & 0 & 1.20 & 0 & 12.63 \\
    w/o Memory Agent & 23.82 & 3.34 & 15.60 & 22.00 & 9.83 & 48.28 \\
    w/o Reflection Mechanism & 25.44 & 3.01 & 17.20 & 23.20 & 8.60 & 45.45 \\
    MA-CoNav (Full) & 33.56 & 2.93 & 25.60 & 37.60 & 15.62 & 69.89 \\
    \bottomrule
  \end{tabular}
  }
\end{table}

\subsubsection{Component-level Ablation Experiments}
To further explore the rationality of the internal design of each sub-agent, we ablated their key components while keeping the framework intact.

% \subsubsubsection{Ablation of Internal Components of Task Planning Agent}
% 对任务规划智能体内部的各个组件进行消融，分析其对性能的影响。结果如表\ref{tab:planning_agent_ablation}所示。禁用任务动态校验后，失去了对每个子任务完成状态的闭环语义评估能力，其任务成功率（SR）降至 15.6\%；移除层级分解模块的影响不如移除动态校验显著，这表明层级分解必须与动态校验结合使用。
We conducted an ablation study on the internal components of the Task Planning Agent to analyze their impact on performance. The results are shown in Table~\ref{tab:planning_ablation}. Disabling the task dynamics validation leads to the loss of closed-loop semantic evaluation capability for the completion status of each subtask, resulting in a drop in success rate (SR) to $15.6\%$. The impact of removing the hierarchical decomposition module is less significant than that of removing dynamics validation, indicating that hierarchical decomposition must be used in conjunction with dynamics validation.

\begin{table}[htbp]
  \centering
  \caption{Ablation Experiment of Internal Components of Task Planning Agent}
  \label{tab:planning_ablation}
  \resizebox{\linewidth}{!}{%
  \begin{tabular}{lcccccc}
    \toprule
    Method & NL (Steps) $\downarrow$ & NE (Meters) $\downarrow$ & SR (\%) $\uparrow$ & OSR (\%) $\uparrow$ & SPL (\%) $\uparrow$ & KPA (\%) $\uparrow$ \\
    \midrule
    w/o Hierarchical Decomposition & 30.35 & 3.65 & 18.40 & 23.60 & 11.04 & 42.42 \\
    w/o Dynamic Verification & 28.19 & 3.91 & 15.60 & 18.80 & 8.11 & 39.86 \\
    MA-CoNav (Full) & 33.56 & 2.93 & 25.60 & 37.60 & 15.62 & 69.89 \\
    \bottomrule
  \end{tabular}
  }
\end{table}

% \subsubsubsection{Ablation of Internal Components of Control Execution Agent}
% 对控制执行智能体内部的各个组件进行消融，分析其对性能的影响。结果如表\ref{tab:execution_agent_ablation}所示。禁用几何地图构建导致SR=9.6\%和最大的终点误差（NE=5.81米），表明缺乏基于坐标的精确位置追踪，机器人无法在物理空间中建立可靠的位姿估计。移除拓扑地图构建后，系统仍能通过几何信息进行避障和移动，但失去了对环境连通性的抽象表示。这导致路径规划变得短视和低效，机器人可能反复探索已访问区域，表现为导航路径（NL）异常增长。
We conducted an ablation study on the internal components of the Control Execution Agent to analyze their impact on performance. The results are shown in Table~\ref{tab:execution_ablation}. Disabling geometric map construction results in an SR of $9.6\%$ and the largest navigation error (NE = $5.81$\,m), indicating that without precise coordinate-based position tracking, the robot is unable to establish reliable pose estimation in physical space. Removing topological map construction still allows the system to perform obstacle avoidance and movement using geometric information; however, it loses the abstract representation of environmental connectivity. This leads to myopic and inefficient path planning, causing the robot to repeatedly explore already visited areas, which manifests as an abnormal increase in navigation length (NL).

\begin{table}[htbp]
  \centering
  \caption{Ablation Experiment Analysis of Internal Components of Control Execution Agent}
  \label{tab:execution_ablation}
  \resizebox{\linewidth}{!}{%
  \begin{tabular}{lcccccc}
    \toprule
    Method & NL (Steps) $\downarrow$ & NE (Meters) $\downarrow$ & SR (\%) $\uparrow$ & OSR (\%) $\uparrow$ & SPL (\%) $\uparrow$ & KPA (\%) $\uparrow$ \\
    \midrule
    w/o Geometric Map Construction & 29.5 & 5.81 & 9.60 & 14.40 & 6.72 & 32.53 \\
    w/o Topological Map Construction & 35.21 & 3.80 & 16.40 & 21.20 & 10.66 & 45.05 \\
    MA-CoNav (Full) & 33.56 & 2.93 & 25.60 & 37.60 & 15.62 & 69.89 \\
    \bottomrule
  \end{tabular}
  }
\end{table}

% \subsubsection{Ablation of Internal Components of Memory Agent}
% 记忆智能体承担记录与学习双重职责，对各个组件进行消融，结果如表\ref{tab:memory_agent_ablation}所示。仅记录原始历史，不进行结构化提炼，相当于拥有一个庞大的“日志库”而非“知识库”，信息冗杂，对当前决策产生不利影响，导致所有性能指标下降；保留经验提炼但禁用检索应用，意味着系统无法在需要时调用历史经验，这导致经验库无法对在线决策产生直接影响。
The memory agent assumes the dual responsibilities of recording and learning, and ablation experiments are conducted on each component, with the results shown in Table~\ref{tab:memory_ablation}.
Recording only the raw history without structured abstraction is equivalent to maintaining a massive ``log library'' rather than a ``knowledge base''.
This redundant information adversely affects the current decision-making process, resulting in the degradation of all performance metrics.
Retaining experience abstraction but disabled the retrieval application implies that the system cannot invoke historical experiences when necessary, which prevents the experience library from exerting a direct impact on online decision-making.

\begin{table}[htbp]
  \centering
  \caption{Ablation Experiment Analysis of Internal Components of Memory Agent}
  \label{tab:memory_ablation}
  \resizebox{\linewidth}{!}{%
  \begin{tabular}{lcccccc}
    \toprule
    Method & NL (Steps) $\downarrow$ & NE (Meters) $\downarrow$ & SR (\%) $\uparrow$ & OSR (\%) $\uparrow$ & SPL (\%) $\uparrow$ & KPA (\%) $\uparrow$ \\
    \midrule
    w/o Experience Structured Refinement & 25.16 & 3.98 & 18.80 & 24.40 & 11.84 & 50.90 \\
    w/o Experience Retrieval and Application & 23.82 & 4.34 & 19.20 & 22.80 & 11.71 & 48.28 \\
    MA-CoNav (Full) & 33.56 & 2.93 & 25.60 & 37.60 & 15.62 & 69.89 \\
    \bottomrule
  \end{tabular}
  }
\end{table}

% \subsubsubsection{Model Selection Experiment for Observation Agent}
% 观察智能体将第一人称视觉流转化为任务导向、富含空间关系的结构化描述。本文评估当前顶尖的不同多模态模型的性能差异，结果如表\ref{tab:model_selection}所示。在障碍物评估方面，GPT-5.2 Pro比Gemini 3 Pro Preview高出5.5\%；尽管Gemini 3 Pro Preview的地标识别准确率达到66.1\%，超过了GPT-5.2 Pro的58.6\%，但在实际应用中其幻觉问题更为严重。鉴于导航任务对空间关系与地标识别的双重高要求，本文最终选择GPT-5.2 Pro作为观察智能体的多模态大模型。
The Observation Agent transforms the first-person visual stream into task-oriented structured descriptions rich in spatial relationships. This paper evaluates the performance differences among current state-of-the-art multimodal models, with results shown in Table~\ref{tab:vision_model_ablation}. In terms of obstacle assessment, GPT-5.2 Pro outperforms Gemini 3 Pro Preview by $5.5\%$. Although Gemini 3 Pro Preview achieves a landmark recognition accuracy of $66.1\%$, surpassing GPT-5.2 Pro's $58.6\%$, its hallucination issue is more severe in practical applications. Given the dual high demands of navigation tasks on both spatial relationships and landmark recognition, this paper ultimately selects GPT-5.2 Pro as the multimodal large model for the Observation Agent.

\begin{table}[htbp]
  \centering
  \caption{Additional Model Selection Experiment}
  \label{tab:vision_model_ablation}
  \resizebox{\linewidth}{!}{%
  \begin{tabular}{lcc}
    \toprule
    Model Name & Obstacle Detection Accuracy (\%) & Landmark Recognition Accuracy (\%) \\
    \midrule
    GPT-5.2 Pro & 79.52 & 58.62 \\
    Gemini 3 Pro Preview & 74.00 & 66.14 \\
    \bottomrule
  \end{tabular}
  }
\end{table}

\subsubsection{Ablation Experiment of Reflection Mechanism}
To quantitatively verify the proposed local-global dual-stage reflection mechanism, three targeted indicators were introduced to measure the local reflection mechanism. Error Detection Rate (EDR): the proportion of steps where local reflection detects potential problems to the total steps. Reflection Accuracy (RA): the proportion of steps judged as risky by the reflection mechanism that are confirmed to have real problems manually afterwards. Rollback Success Rate (RSR): the proportion of subtasks that are finally successful after triggering local reflection and performing action rollback/adjustment.

In complex long-instruction tasks, the error detection rate is as high as 18.7\%, indicating that the system frequently faces inconsistent perception and planning, highlighting the necessity of real-time verification. At the same time, the reflection accuracy (RA) of 85.2\% indicates that the mechanism has high judgment accuracy, effectively reducing unnecessary interventions. Most importantly, 72.3\% of all cases that triggered reflection and adjusted strategies finally completed the current subtask successfully.

% \begin{table}[htbp]
%   \centering
%   \caption{Effectiveness Analysis of Local Reflection Mechanism}
%   \label{tab:reflection_ablation}
%   \resizebox{0.6\linewidth}{!}{%
%   \begin{tabular}{lccc}
%     \toprule
%     Method & DER (\%) $\uparrow$ & RA (\%) $\uparrow$ & RSR (\%) $\uparrow$ \\
%     \midrule
%     w/o Local Reflection & 0\% & -- & -- \\
%     Local Reflection & 18.7\% & 85.2\% & 72.3\% \\
%     \bottomrule
%   \end{tabular}
%   }
% \end{table}

To evaluate the accuracy of the retrieval module after global reflection, the "Memory Retrieval Accuracy (MRA)" indicator was introduced: the proportion of cases where the top-ranked case returned is truly relevant to the current scene among all retrieval requests initiated by the system. Experimental results show that MRA reaches 78.4\%. This means that in most cases, when the system encounters a new scene, the memory agent can accurately recall the most relevant historical experience from the case library.

On the basis of ensuring retrieval quality, we further quantified the actual application effect of the recalled experience. After accumulating experience from about 50 initial tasks, the system was tested on a new set of 20 tasks. Among the new tasks, the proportion of tasks where the system successfully retrieved relevant historical experience from the memory library (scene matching degree > threshold) was 35\%.

\section{Conclusion}
% 本文提出了MA-CoNav，一种全新的主从式多智能体协同导航框架。该框架采取“主-子”层级协作模型，将导航所必需的全局认知能力，解耦并分配给四个各司其职的专业化智能体。该框架的导航性能显著优于单智能体和多智能体基准方法。大量实验表明，MA-CoNav 适用于复杂长程任务，能够高效准确地解析、规划并执行任务。而内置的反思机制不仅增强系统的稳健性，还提高误差修正效率，特别是在较长的导航任务中，成功率显著提升。未来，我们的工作将聚焦于如何融合多传感器信息以提升在动态、非结构化环境中的感知鲁棒性与避障能力以及如何将外部常识知识库与智能体的内部长期记忆相结合，使其能够理解和执行需要外部世界知识或历史情境的复杂指令（如“把文件放到上次开会用的桌子上”）。

This paper proposes MA-CoNav, a novel master-slave multi-agent collaborative navigation framework. This framework adopts a \textit{master-slave} hierarchical collaboration model, which decouples and distributes the global cognitive capabilities necessary for navigation to four specialized agents with dedicated responsibilities. The navigation performance of this framework is significantly superior to that of single-agent and multi-agent baseline methods. Extensive experiments demonstrate that MA-CoNav is applicable to complex long-haul tasks and can parse, plan and execute tasks efficiently and accurately. The built-in reflection mechanism not only enhances the robustness of the system but also improves the efficiency of error correction; in particular, the success rate is significantly improved in long navigation tasks. For future work, we will focus on how to fuse multi-sensor information to improve perceptual robustness and obstacle avoidance capabilities in dynamic, unstructured environments, and how to combine external commonsense knowledge bases with the agents' internal long-term memory to enable them to understand and execute complex instructions that require external world knowledge or historical context.





%% For numbered reference style

%   \textit{\LaTeX: a document preparation system},
%   Addison Wesley, Massachusetts,

% \bibliographystyle{elsarticle-num}   % 参考文献样式
\bibliography{references}

\end{document}